\documentclass[10pt,twocolumn,letterpaper]{article}

\usepackage[pagenumbers]{packages/wacv} 

\usepackage{graphicx}
\usepackage{amsmath}
\usepackage{amssymb}
\usepackage{booktabs}
\usepackage[accsupp]{axessibility}

\usepackage{gensymb}
\usepackage{booktabs, tabularx}
\usepackage{caption}

\usepackage{acro}

\usepackage[pagebackref,breaklinks,colorlinks]{hyperref}
\DeclareTextSymbolDefault{\dh}{T1}
\usepackage[capitalize]{cleveref}
\crefname{section}{Sec.}{Secs.}
\Crefname{section}{Section}{Sections}
\Crefname{table}{Table}{Tables}
\crefname{table}{Tab.}{Tabs.}

\def\wacvPaperID{621}
\def\confName{WACV}
\def\confYear{2025}

\DeclareAcronym{cvgl}{
    short = CVGL,
    long  = Cross-View Geo-localisation,
    tag   = nomencl
}
\DeclareAcronym{fov}{
    short = FOV,
    long  = Field-of-View,
    tag   = nomencl
}
\DeclareAcronym{bev}{
    short = BEV,
    long  = Birds-Eye-View,
    tag   = nomencl
}
\DeclareAcronym{gnss}{
    short = GNSS,
    long  = Global Navigation Satellite Systems,
    tag   = nomencl
}
\DeclareAcronym{sota}{
    short = SOTA,
    long  = state of the art,
    tag   = nomencl
}
\DeclareAcronym{bvm}{
    short = BVM,
    long  = Bearing Vector Matching,
    tag   = nomencl
}

\DeclareAcronym{paper_name}{
    short = SpaGBOL,
    long  = Spatial-Graph-Based Orientated Localisation,
    tag   = spagbol
}
\DeclareAcronym{gis}{
    short = GIS,
    long  = Geographic Information Systems,
    tag   = gis
}

\newlength{\oldtabcolsep}
\begin{document}
\title{SpaGBOL: Spatial-Graph-Based Orientated Localisation}

\author{
    $\text{Tavis Shore}^1$ \qquad
    $\text{Oscar Mendez}^2$ \qquad
    $\text{Simon Hadfield}^1$ \\
    $\text{University of Surrey}^1$ \qquad
    $\text{Locus Robotics}^2$ \\
    {\tt\small \{t.shore, s.hadfield\}@surrey.ac.uk}, 
    {\tt\small omendez@locusrobotics.com} \\
}

\maketitle

\begin{abstract}
Cross-View Geo-Localisation within urban regions is challenging in part due to the lack of geo-spatial structuring within current datasets and techniques. 
We propose utilising graph representations to model sequences of local observations and the connectivity of the target location.
Modelling as a graph enables generating previously unseen sequences by sampling with new parameter configurations.
To leverage this newly available information, we propose a GNN-based architecture, producing spatially strong embeddings and improving discriminability over isolated image embeddings.
We outline SpaGBOL, introducing three novel contributions. 
1) The first graph-structured dataset for Cross-View Geo-Localisation, containing multiple streetview images per node to improve generalisation. 
2) Introducing GNNs to the problem, we develop the first system that exploits the correlation between node proximity and feature similarity. 
3) Leveraging the unique properties of the graph representation - we demonstrate a novel retrieval filtering approach based on neighbourhood bearings.
SpaGBOL achieves state-of-the-art accuracies on the unseen test graph - with relative Top-1 retrieval improvements on previous techniques of 11\%, and 50\% when filtering with Bearing Vector Matching on the SpaGBOL dataset. Code and dataset available: \href{https://github.com/tavisshore/SpaGBOL}{github.com/tavisshore/SpaGBOL}.
\end{abstract}

\vspace{-1em}

\begin{figure}[t!]
    \centering
    \includegraphics[width=\columnwidth]{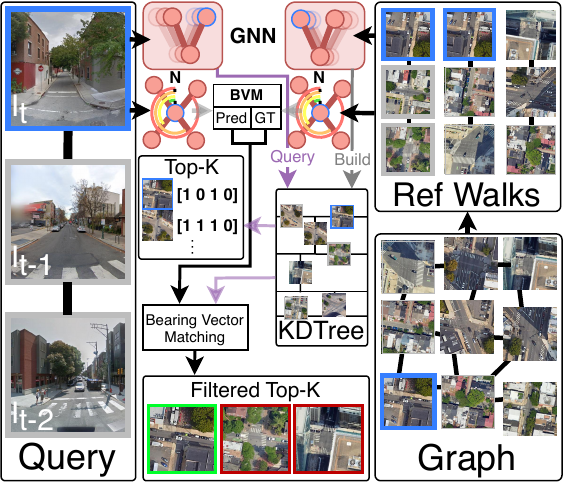}
    \caption{
    At inference time, a KDTree is constructed from exhaustive reference walks sampled from the city's graph. 
    A randomly selected query walk passes through the network, retrieving corresponding embeddings from the KDTree ordered in descending similarity.
    These are further filtered to the set of compatible nodes with \ac{bvm}.
    }
    \label{fig:front}
    \vspace{-1em}
\end{figure}

\section{Introduction}
Localisation is essential in many robotics applications.
Techniques like \ac{gnss} provide absolute positioning data but often fail in environments like urban canyons, where occlusions and reflections interfere with satellite signals.
Image-based localisation offers an alternative approach, enabling a machine to determine its position by capturing images of its surroundings and comparing them to pre-recorded geo-referenced images.
Most modern vehicles are equipped with cameras, simplifying the adoption of image-based localisation. 

\vspace{0.4em}

Two main retrieval-based image localisation techniques are: image-to-image localisation, where query and reference images are taken from the same perspective, and \ac{cvgl}, where street view query images are matched with a database of satellite images.
Both with the same objective - returning the geographic coordinates of the retrieved image.
Existing \ac{cvgl} techniques primarily focus on sparse streetview-satellite image pairs - randomly sampled from across vast regions, disregarding the geo-spatial structure and relationships between neighbouring regions.
Sequential \ac{cvgl} extends single-image techniques, querying multiple images to strength representations - extracting features with cross-frame information. This provides a more practical solution, and estimates position with higher confidence and precision.
These datasets and techniques succeed in learning related features between the viewpoints but still consider data as sequences of separate image pairs with no spatial structure beyond chronology. 
Reference data remains unstructured with no geo-spatial metadata, limiting real-world representational accuracy.
This can make it challenging to recognise new sequences which partially overlap or combine several existing sequences seen during training.
To improve the feasibility of \ac{cvgl}, research should be focused to regions most likely to experience \ac{gnss} communication failure, dense urban city centres. 
The design of image localisation techniques should progress to expect any possible sequence of images within the considered regions. 

\vspace{0.4em}

We propose structuring image localisation data as graph networks. 
This adds crucial geo-spatial information, enabling the generation of unseen sequences of desired length. 
Progressing to this data representation is relatively simple as the target of our system, urban canyons within dense city centres, generally have existing accurate graph representations within many \ac{gis}. 
We therefore propose utilising GNNs to improve \ac{cvgl} within this novel representation, storing sets of streetview images and satellite images at junctions (graph nodes), with connecting roads represented as the graph edges between these nodes. A brief overview of the proposed system is displayed in Figure \ref{fig:front}.
To solidify our proposal into the progression of \ac{cvgl} towards real-world feasibility, we release the \textit{\ac{paper_name}} dataset: a dense multi-city graph-based \ac{cvgl} dataset with multiple streetview images per satellite image - allowing for generalisation across time, weather, and lighting. 
This dataset is split into training and test sets, comprising of 9 cities and 1 city respectively. 
We prove the positive impact that graph representation has on \ac{cvgl} performance due to strengthened feature representation and filtering by neighbourhood road bearings - valid within this city-scale due to neighbouring node's close proximity. \\

\noindent
In summary, our research contributions are:
\begin{itemize}
    \item Introduce a new direction for \ac{cvgl} research, moving from sparse cross-view image retrieval and sequential image retrieval into spatially-strong dense image retrieval, moving the field closer to real-world feasibility for assisting \ac{gnss} techniques in urban environments.
    \item Propose an introductory GNN model utilising data along graph walks to create strong representations, also exploiting derived characteristics to filter retrievals with \ac{bvm}, greatly improving performance.
    \item Release a dense multi-city graph-based \ac{cvgl} dataset, \textit{\ac{paper_name}}, containing train and test set graphs with corresponding images from a sample of the densest city centres across the globe. 
\end{itemize}

\section{Related Works}
\subsection{Cross-View Geo-Localisation} 
The predominant technique for \ac{cvgl} is embedding retrieval. Novel techniques are being proposed at an increasing rate, aiming to improve performance by manipulating extracted features, \cite{Zhu2022TransGeoTI}, \cite{SAIG}, \cite{shore2023bevcv}. 

Deep learning was first introduced to \ac{cvgl} by Workman and Jacobs \cite{7301385}, utilising CNNs for correlated feature extraction across viewpoints, proving their suitability. Lin et al. \cite{7299135} extended this by regarding each query as unique - using euclidean similarities for retrieving clusters. Vo and Hays \cite{Vo2016LocalizingAO} then utilised aerial rotational information with an auxiliary loss, observing the impact of image misalignment - leading to our incorporating of a compass in order to aid system performance. CVM-Net \cite{8578856} appended NetVLAD \cite{netvlad} to a siamese CNN architecture, aggregating residuals of local features to cluster centroids - improving accuracy though greatly increasing complexity.  Zhu et al. \cite{Zhu2020RevisitingSV} leveraged activation maps to estimate orientation. Sun et al. \cite{Sun2019GEOCAPSNETGT} created a capsule network following a ResNet backbone, improving upon CVM-Net performance by approximately 10\%. Liu and Li \cite{8954224} inserted orientation information to the problem, improving the representational robustness of their latent space. 
\begin{figure*}[t!]
    \centering
    \includegraphics[width=0.95\textwidth]{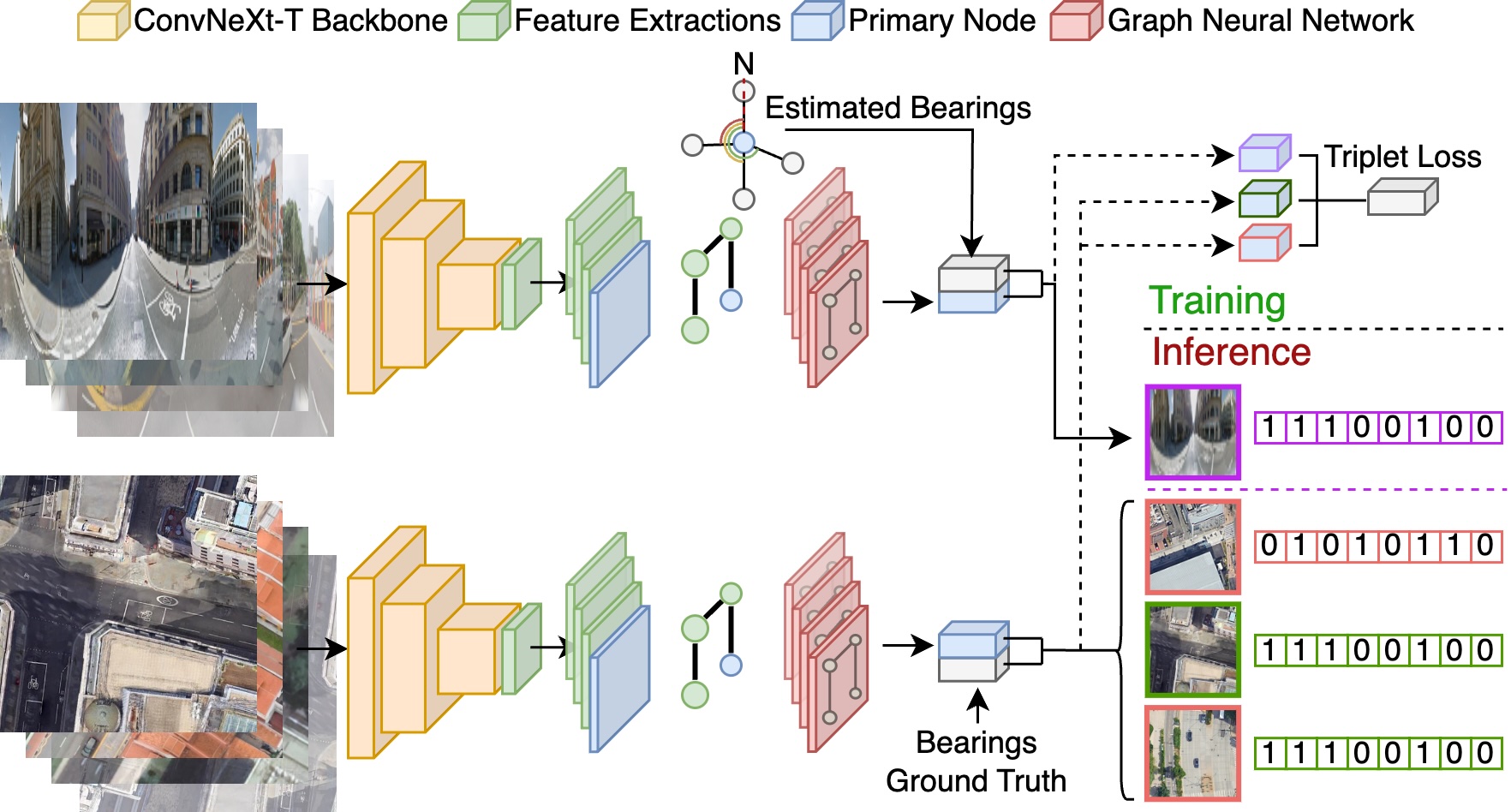}
    \caption{\ac{paper_name} is a two-branch neural network with no weight-sharing, from left to right the network performs the following actions: (1) Image feature extraction with ConvNext-T, (2) Depth-first walk image features $\rightarrow$ GNN embedding (red), (3) Produce neighbour bearing vectors, (4) Perform embedding retrieval from the KDTree, (5) Filter retrievals with bearings to return final geo-coordinates.}
    \label{fig:network}
    \vspace{-1em}
\end{figure*}
Shi et al. \cite{Shi2019SpatialAwareFA} developed a spatial attention mechanism, improving feature alignment between views. Regmi et al. \cite{Regmi2019BridgingTD} created a conditional GAN to synthesise aerial representations of ground-level panoramas. Shi et al. \cite{Shi2019OptimalFT}, \cite{Shi2020WhereAI} proposed techniques for increasing the similarity of features across viewpoints before applying them to limited-\ac{fov} data. This is important due to the ubiquity of monocular cameras compared with panoramic cameras; essential for wide-spread feasibility and adoption. \cite{Shi2020WhereAI} computes feature correlation between ground-level images and polar-transformed aerial images, shifting and cropping at the strongest alignment before performing image retrieval.

Toker et al. \cite{Toker2021ComingDT} proposed synthesising streetview images from aerial image queries before performing image retrieval. L2LTR \cite{Yang2021CrossviewGW} developed a CNN+Transformer network, combining a ResNet backbone with a vanilla ViT encoder. TransGeo \cite{Zhu2022TransGeoTI} proposed a transformer that uses an attention-guided non-uniform cropping strategy to remove uninformative areas.
In GeoDTR \cite{GeoDTR} and their following work GeoDTR+ \cite{GeoDTR+}, Zhang et al. disentangle geometric information from raw features, learning spatial correlations among visual features to increase performance.
Zhu et al. \cite{SAIG} introduce \textit{SAIG}, an attention-based backbone for \ac{cvgl}, representing long-range interactions among patches and cross-view relationships with multi-head self-attention layers.
BEV-CV \cite{shore2023bevcv} introduces \ac{bev} transforms, further reducing the representation difference between viewpoints to create more similar embeddings. Sample4Geo \cite{sample4geo} propose two sampling strategies for \ac{cvgl}, sampling geographically for optimal training initialisation, and mining hard-negatives according to visual similarities between embeddings. 
Generally, the above works all focus on developing more similar embeddings for either sparsely sampled image pairs or relatively limited image sequences. In contrast, we transition \ac{cvgl} to methods that more closely represent real \ac{gnss}-denied regions, advancing the field towards practical application.

\subsection{Graph-Based Localisation}
Graph-networks and GNNs have not previously been utilised in the field of \ac{cvgl}. They have however been applied to related fields, from localising objects within scene graphs to mapping out environments for graph-based SLAM. We outline some key related works that contributed to our proposition of their application for \ac{cvgl}.

Graph-based SLAM techniques construct a graph mapping of an environment while simultaneously localising an agent within the map. Heinzle et al. \cite{heinzle2005graph} introduce pattern recognition within road networks - aiming to perform automatic localisation of city centres. Grisetti et al. \cite{5681215} display an overview of Graph-based SLAM methods, representing generally GNSS-denied indoor environments as graphs, localising within the graph using probabilistic techniques. Kümmerle et al. \cite{kummerle2011large} introduces the use of aerial priors alongside sensor data to improve map creation for graph-based SLAM. Annaiyan et al. \cite{7991524} use stereo imaging to construct and localise UAVs within a graph-based map. He et al. \cite{9234707} combine visual-LIDAR data to construct 3D maps of environments, merging with a pose graph optimisation procedure. Vysotska and Stachniss \cite{vysotska2016lazy} present a search heuristic aiming to efficiently find matches between an image sequence and a database using a data association graph.
Johnson et al. \cite{7298990} introduce a framework for semantic image retrieval based on scene graphs, outperforming methods that only use low-level image features. Liu et al. \cite{8794475} leverage object level semantics and spatial environment understanding for localisation, improving performance where extreme appearance changes occur. Giuliari et al. \cite{Giuliari_2022_CVPR} use Spatial Commonsense Graphs to localise objects in partial scenes where nodes represent objects, and edges represent pairwise distances between them.
Finally we outlined examples of practical applications of both graph structures and GNNs. \cite{ORNGARDARSSON2022661} represent water utility networks as graphs, using Graph Convolutional Networks (GCNs) to predict nodal pressures, and localise leaks. In a similar manner, \cite{GRATTAROLA2022117330} introduce graphs and GNNs to localise epileptic seizure onset zones, where nodes represent different regions of the brain. Murai et al. \cite{10286058} developed a graph-based collaborative localisation system for robots, globally localising via efficient peer-to-peer communication. 
Most prior graph and GNN works have attempted to learn similarities between related examples from the same domain. In our work we attempt to preform cross-view graph matching between images on the ground, and those from a satellite.

\section{Methodology}
\begin{figure}[t!]
  \centering
    \includegraphics[width=\columnwidth]{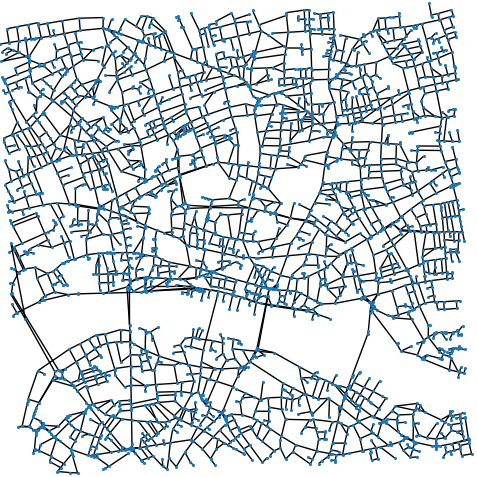}
    \caption{Corpus graph of London City Centre. Each graph is square with sides of length 2km. Nodes (junctions) are shown here in blue, with black edges (roads).}
    \label{fig:london_graph}
    \vspace{-1.5em}
\end{figure}

\subsection{CVGL Graph Representation}
To store geographically dense collections of images with a strong spatial structure we propose a graph representation, improving feasibility and extending the potential techniques suitable for \ac{cvgl} - an example graph is shown in Figure \ref{fig:london_graph}. 
We represent cities $i \in \{London, Tokyo,...\}$ as separate graphs $G_i=(N_i, E_i)$ with nodes $N_i = \{n_1, n_2, ..., n_N\}$ and edges $E_i = \{e_{1,2}, e_{1,3}, ..., e_E \}$. 
Nodes $n$ represent road junctions and edges $e_{a,b}$ represent roads connecting nodes ${a}$ and ${b}$. Figure \ref{fig:graph_splits} shows how the graphs are separated into train/validation/test sets.
For each node we collect a satellite image and 5 corresponding panoramic streetview images captured over an extended period. 
Both image types are RGB: $I_{t} \in \mathbb{R}^{3{\times}W{\times}H}, t \in \{street, sat\}$. Each node holds attributes - $n_i = \{I_{sat}, I_{street}^{1..5}, L, \Psi, B \}$, where location $L=\{\phi, \lambda\}$ contains geographical latitude and longitude coordinates, $\Psi \in \mathbb{R}: \{-180\degree \leq \Psi \leq 180\degree\}$ is the north-centred camera yaw, and $B=\{\beta_{1}, ..., \beta_{K}\}$ are north-aligned bearings to it's $K$ neighbouring nodes - where $\beta \in \mathbb{R}: \{-180\degree \leq \beta \leq 180\degree\}$.

The panoramic streetview image ($I_{street}^*$) \ac{fov} is varied to evaluate the feasibility of using monocular cameras. Cameras are assumed to be fixed to the vehicle in a forward-facing configuration. Where \ac{fov}, $\Theta \in \{360\degree, 180\degree, 90\degree\}$:

\vspace{-0.5em}

\begin{equation}
I_{street} = \mathrm{fov\_crop}\left(I_{street}^*, \Theta, \Psi\right)\,
\label{eq:crop}
\end{equation}

The proposed system takes randomly sampled query walks (exhaustive for reference set) $W_{i}^{j}$ of length $l \in \{1, ..., 5\}$ as input from each node $n_j$ in graph $G_i$
\begin{equation}
    W_{i}^{j} = \mathrm{random\_walk}(G_{i}(n_{j})) 
    \label{eq:sam}
\end{equation}

\begin{figure}[t]
  \centering
    \includegraphics[width=0.95\columnwidth]{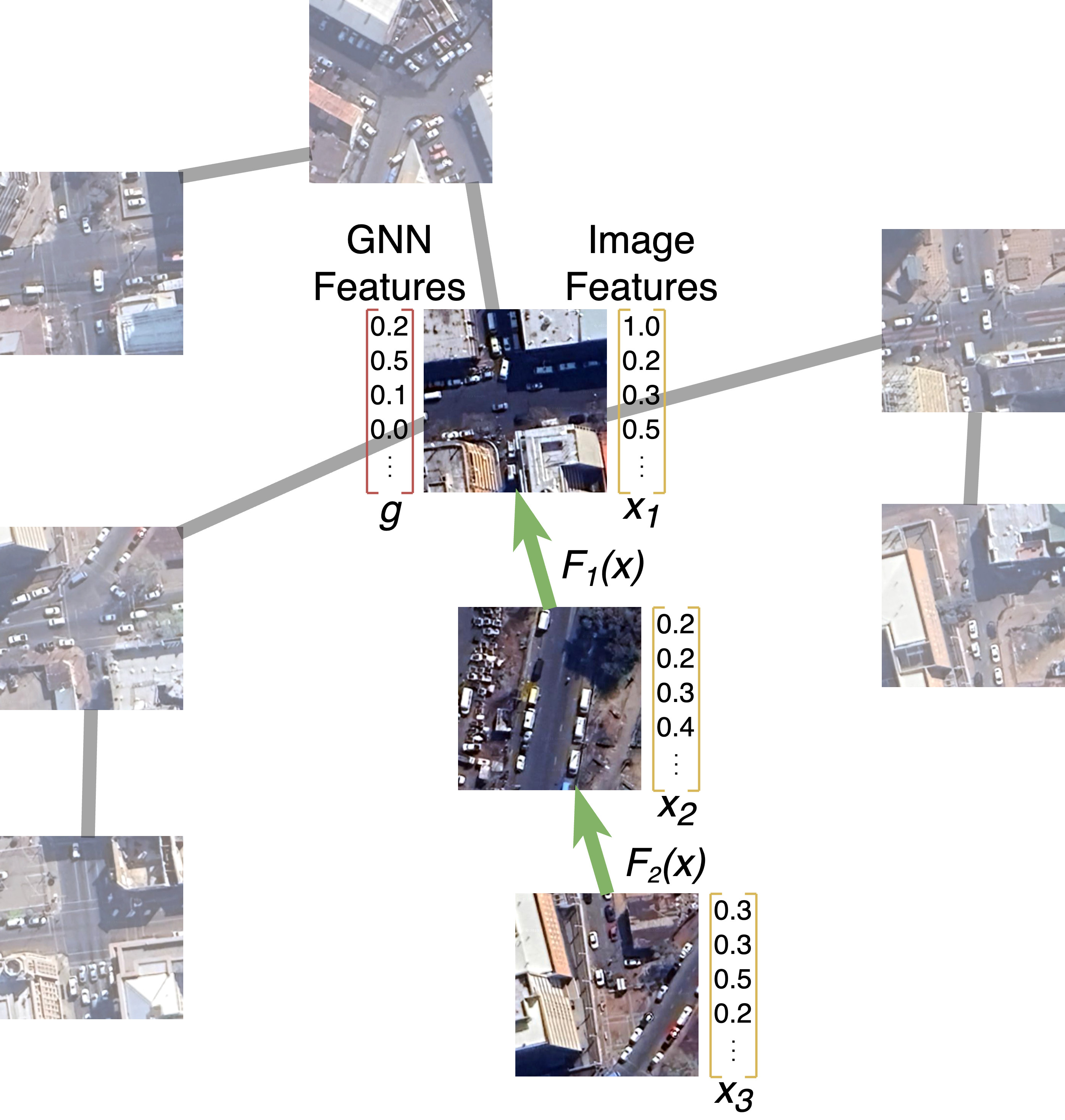}
    \caption{Random depth-first walk sample of length 3. Image features are extracted from each node, passing through a GNN to produce the final node embedding.}
    \label{fig:walk_sample}
    \vspace{-1.5em}
\end{figure}

A walk representation is shown in Figure \ref{fig:walk_sample}, randomly selecting one depth-first walk from the target node's available walks. This walk is then extracted from the corpus graph as a subgraph - passing the streetview images, satellite images, and other attributes through the corresponding branches within the \ac{paper_name} network. 
The training/validation/testing walks are sampled from disconnected graphs and subgraphs, as shown in Figure \ref{fig:graph_splits}. 

\subsection{\ac{paper_name} Neural Network}
During training, corresponding streetview and satellite image walks are passed through \textit{\ac{paper_name}}, shown in Figure \ref{fig:network}. The network's upper and lower branches are identical but do not share any weights. Streetview queries are passed through the upper branch and corresponding satellite targets through the lower branch. Each branch first embeds it's inputs through CNN backbones:

\begin{figure}[t]
    \includegraphics[width=\columnwidth]{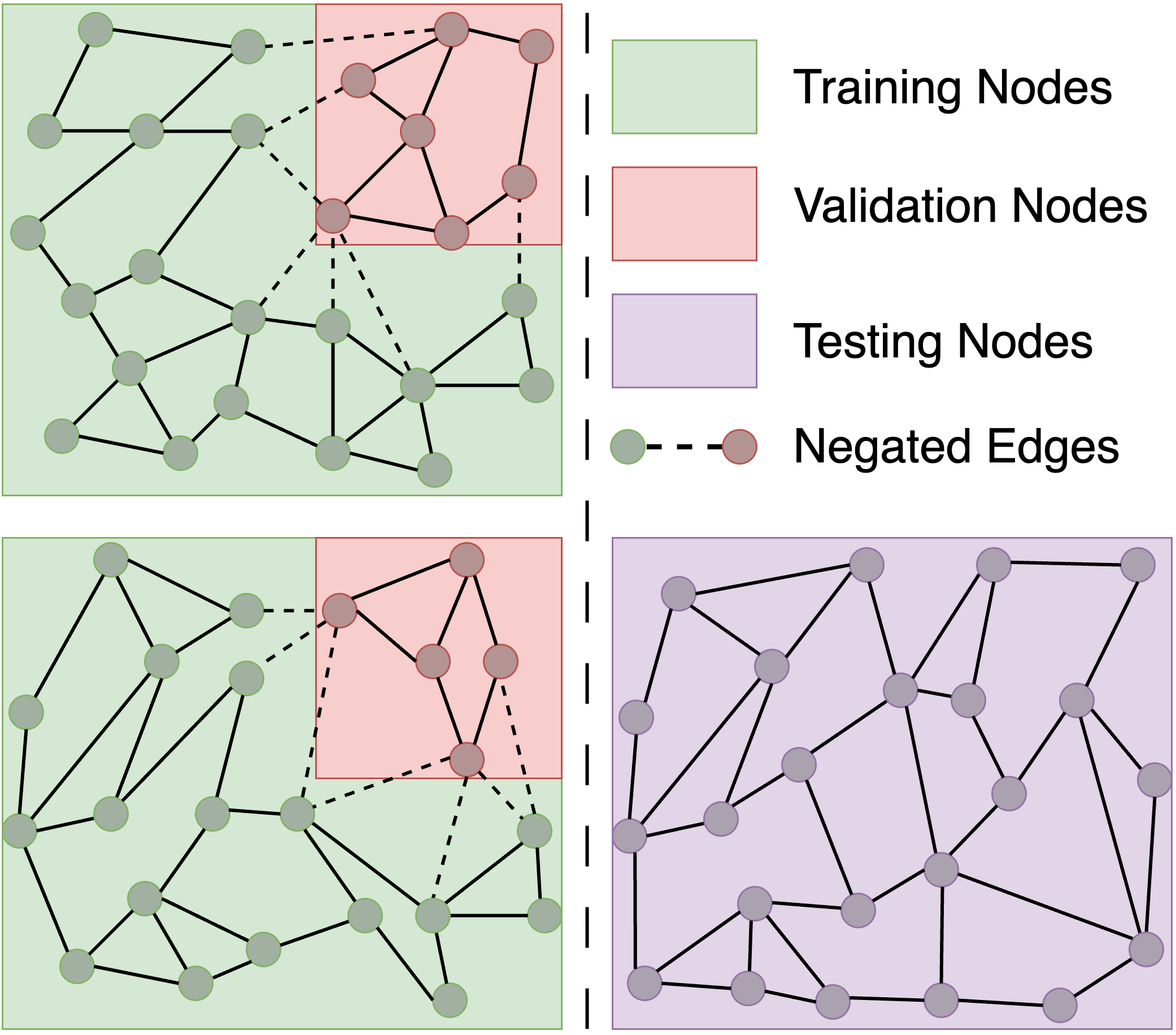}
    \caption{Splitting corpus graphs into train/validation/test sets. Validation graphs are unconnected subgraphs of each training graphs. The test graph is a wholly unseen city graph.}
    \label{fig:graph_splits}
    \vspace{-1em}
\end{figure}

\begin{equation}
    feat_{street} = \mathrm{CNN_{street}}\left(I_{street}^{rand(0-4)}\right)
    \label{eq:feat_street}
\end{equation} 

\vspace{-0.5em}

\begin{equation}
    feat_{sat} = \mathrm{CNN_{sat}}\left(I_{sat}\right).
    \label{eq:feat_sat}
\end{equation}

A sequence of GNN layers then process the results, as
\begin{equation}
h_{n_{j}}^{k+1} = \sigma \left( \Omega^k \cdot \mathrm{AGG}\left(\{ h_{n_{u}}^k, \forall u \in W_F \}\right) \right)
\label{eq:gnn}
\end{equation}
where $h_{n_{j}}^{k+1}$ is the updated embedding of node, $n_{j}$ at layer $k+1$, $\sigma$ is an activation function, $\Omega^k$ is a weight matrix for layer $k$, $\mathrm{AGG}$ is a mean-based aggregating function combining features from neighbouring nodes, $h_{n_{u}}^k$ is the embedding of node $n_{u}$ at layer $k$, $W_{F}$ is the set of walk image features where $F \in \{feat_{street}, feat_{sat}\}$. The output graph embedding from the final layer is then $h_{n_{j}}^L$. For the streetview branch these final embeddings are notated as  $\eta_{street}^{j}$ while the satellite branch embeddings are $\eta_{sat}^{j}$. 

The network is trained using a triplet loss function, with the objective of producing similar GNN embeddings for corresponding streetview and satellite walks. We select walk triplets by deeming a walk of streetview images as the anchor, it's corresponding walk of satellite images as the positive, and randomly selecting an unrelated walk of satellite images as the negative. 
More specifically, we utilise the Triplet Loss Function:

\vspace{-0.5em}

\begin{equation} 
\mathcal{L}\!=\!\sum_{i=1}^{N} \left[ \left\| \eta_{street}^{a}-\eta_{sat}^{p} \right\|_2^2 - \left\| \eta_{street}^{a} - \eta_{sat}^{n} \right\|_2^2 + \alpha \right]
\end{equation}

where \( \eta_{street}^a \), \( \eta_{sat}^p \), and \( \eta_{sat}^n \) are the anchor, positive, and negative embeddings, respectively, \( \left\| \cdot \right\|_2 \) is the Euclidean norm, and \( \alpha \) is the margin.

\begin{figure}[!ht]
    \centering
    \includegraphics[width=\columnwidth]{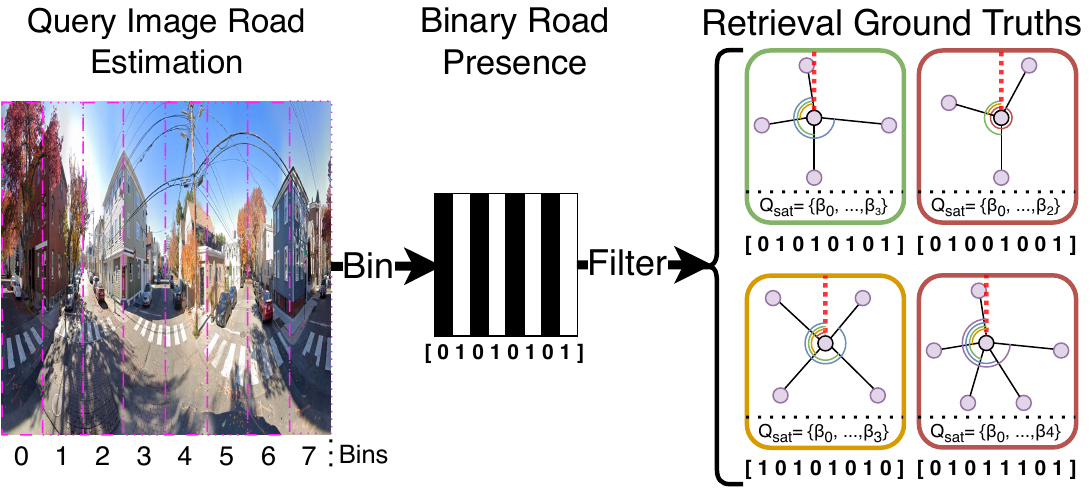}
    \caption{Road bearings may be estimated from panoramic streetview images into a configurable number of bins. These can then be matched against quantised bearings for retrievals.}
    \label{fig:bearings}
    \vspace{-1em}
\end{figure}

\subsection{Bearing Vector Matching}
A significant benefit of utilising graphs for \ac{cvgl} is the ability to efficiently filter route proposals.
We pre-compute both the number of neighbours at each node, and the relative bearings (azimuth) to each neighbour. These relative bearings $\theta \in \{\beta_{0},..., \beta_{K}\}$ are calculated using the geographic coordinates of the two nodes ($a$ and  $b$):

\vspace{-1em}

\begin{equation}
\!\!\beta_{b}\!=\!\text{acos}\!\left(\sin(\phi_a\!)\!\cdot\!\sin(\phi_b\!)\!+\!\cos(\phi_a\!)\!\cdot\!\cos(\phi_b\!)\!\cdot\!\cos(\Delta\lambda)\right)
    \label{eq:bearing}
\end{equation}

where $\Delta\lambda$ is the difference in longitude and $\phi$ is the latitude of each node.
These bearings are then quantised into $V$ bins, in the bearings vector $Q = \left(Q_0, Q_1,... , Q_V\right)$:

\vspace{-1em}

\begin{equation}
    Q_{\upsilon} = \begin{cases} 
                    1 & \!\!
                    \exists\theta \in \{\beta_{0},..., \beta_{K}\}
                    \text{ such that }\!
                    \frac{\upsilon}{V} \!\!<\!\! \frac{\theta}{2\pi} \!\!\leq\!\! \frac{\upsilon+1}{V} \\
                    0 & \!\! \text{otherwise} 
                \end{cases}
    \label{eq:bin_eq}
\end{equation}

This creates a binary code describing the arrangement of roads at this junction. Bins of equal width are used where bin width $\omega = \frac{V}{360}$ degrees, shifted by $\frac{\omega}{2}$ degrees as the camera is expected to be forward-facing, leaving the forwards road appearing in the centre of the midpoint bin. All reference bearing vectors $Q_{sat} = \{Q_0, ..., Q_N\}$ are computed prior to evaluation. 

At query time, bearings $Q_{street}$ may similarly be estimated from the streetview images. For example a semantic segmentation or \ac{bev} system can recognise areas of road in different directions. The query and reference junction vectors are then used to filter the image retrievals, discarding retrievals with incompatible bearing vectors. More formally, a retrieval is compatible if any bitwise shift operation of the query matches the retrieval. This operation results in filtered reference retrievals $Q_{sat}^*$ from the overall reference set $Q_{sat}$ whose bearing vectors equal the queries at some shift.

\vspace{-1em}

\begin{equation}
\!Q_{sat}^*\!=\left\{
                Q\!\in\!Q_{sat} | 
                \exists\upsilon 
                \text{ such that } 
                Q_{street}\!=\!\mathrm{shift}(Q,\!\upsilon)
                \right\}
\end{equation}

Performance can be further increased if the vehicle's yaw is known. In this case, the input to the shift operation is defined by the yaw. Figure \ref{fig:bearings} illustrates the bearing filtering technique. The right-hand side displays retrievals determined from \ac{paper_name} along with their pre-computed bearing vectors. These are filtered using the query bearings vector, determined from the query image. 
In this example, the red-outlined embeddings are discarded as their vectors don't match the queries. The orange-outlined embedding is a partial match, with the correct road positions but misaligned. The green-outlined embedding shows a perfect match. Once retrievals have been filtered, the potential retrievals can be greatly narrowed down, increasing the probability of a correct localisation.

\section{Results}
\begin{table}[t!]
    \centering
    \begin{tabular}{cccc}
        \multicolumn{4}{c}{\textbf{\ac{paper_name}}} \\ \hline
        \multicolumn{1}{c|}{Region} & Nodes & Edges & Walks \\ \hline
        \multicolumn{1}{c|}{Tokyo} & 4,815 & 7,942 & 95,044 \\
        \multicolumn{1}{c|}{London} & 3,155 & 4,124 & 30,634 \\
        \multicolumn{1}{c|}{Philly} & 2,272 & 3,782 & 47,774 \\
        \multicolumn{1}{c|}{Brussels} & 2,190 & 3,403 & 35,959 \\
        \multicolumn{1}{c|}{Boston} & 1,567 & 2,403 & 26,180 \\
        \multicolumn{1}{c|}{Guildford} & 1,472 & 1,773 & 11,247 \\
        \multicolumn{1}{c|}{Chicago} & 1,159 & 1,935 & 25,824 \\
        \multicolumn{1}{c|}{New York} & 1,103 & 1,983 & 29,668 \\
        \multicolumn{1}{c|}{Singapore} & 1,043 & 1,567 & 15,241 \\
        \multicolumn{1}{c|}{Hong Kong} & 995 & 1,440 & 13,270 \\ \hline
        \multicolumn{1}{c|}{Total} & 19,771 & 28,912 & 330,841\\ \hline
        \multicolumn{1}{c|}{Streetview} & 98,855 &  &  \\
        \multicolumn{1}{c|}{Satellite} & 19,771 &  & \\

    \hline
        \multicolumn{4}{c}{\textbf{VIGOR-Graph}} \\ \hline
        \multicolumn{1}{c|}{Region} & Nodes & Edges & Walks \\ \hline
        \multicolumn{1}{c|}{New York} & 3,880 & 6,771 & 96,176 \\
        \multicolumn{1}{c|}{San Francisco} & 3,288 & 5,337 & 67,942 \\
        \multicolumn{1}{c|}{Seattle} & 3,039 & 4,697 & 51,370 \\
        \multicolumn{1}{c|}{Chicago} & 2,295 & 3,771 & 49,212 \\ \hline
        \multicolumn{1}{c|}{Total} & 12,502 & 20,576 & 264,700
    \end{tabular}
    \caption{Graph Attributes - No. unique walk samples of length 4}
    \label{tab:graph_details}
    \vspace{-0.5em}
\end{table}

\subsection{Datasets} 
The most significant current \ac{cvgl} datasets (CVUSA \cite{cvusa} and CVACT \cite{8954224}) are unsuitable for conversion to a graph structure as the data is too sparse. We convert the older benchmark dataset VIGOR \cite{vigor} into a graph structure, enabling similar assessment. VIGOR contains densely collected image pairs from four cities within the USA: New York, San Francisco, Chicago, and Seattle.
To convert VIGOR to a graph representation, we first retrieve the graphs for each of these cities, with the same characteristics as \ac{paper_name} - nodes represent junctions and edges represent roads. Each node is then assigned the image pairs closest to their geographical coordinates. This results in 10,207 training nodes and 2,295 testing nodes - the system is evaluated with sampled walks in the same manner as with \ac{paper_name}. \ac{paper_name}'s and VIGOR-Graph's characteristics are displayed in Table \ref{tab:graph_details}, with the total number of walks (when walk length $n=4$) to demonstrate the extensive sampling capabilities when using graph structures. 

\ac{paper_name} contains 18,204 Satellite-Streetview training+validation pairs and 1,567 testing pairs from across 10 cities, covering $2km^2$ per city. Satellite images are north-aligned with a resolution of 0.2metres / pixel covering $50m^2$ (note some of these images may have been captured from drones and other aerial image sources). Streetview images are yaw-aligned panoramas with a resolution of $2048\times512$. When limiting the \ac{fov}, images are cropped to the desired \ac{fov} with yaw rotated away from the previous node. We use Boston's graph as the test set, with the remaining cities used for training - separating a ninth of each training graph for validation, as shown in Figure \ref{fig:graph_splits}. More in-depth information about the \ac{paper_name} dataset is given in the Supplementary Material.

\subsection{Implementation Details}
Image features are extracted with a ConvNext-T \cite{liu2022convnet2020s}, producing 768-dimension embeddings. 
The sampled walk embeddings are passed through a GNN which outputs refined 64-dimension embeddings. 
All image embeddings affect network learning, but only the target node embeddings are retained for evaluation. 
A KDTree of satellite image embeddings is constructed. This is then queried with each streetview image to retrieve the $K$ closest embeddings. 
Training occurs end-to-end, randomly sampling walks of length $n$ for each node per epoch, also randomly selecting the streetview image from each node's streetview set. 
\ac{paper_name} is trained with walk triplets for 100 epochs using an AdamW optimiser with an initial learning rate of 1e-4 and a ReduceLROnPlateau scheduler. 
Graphs during validation and testing are distinct subsets, with one random query walk per node and exhaustive reference walks.

\begin{figure}[t]
  \centering
    \includegraphics[width=\columnwidth]{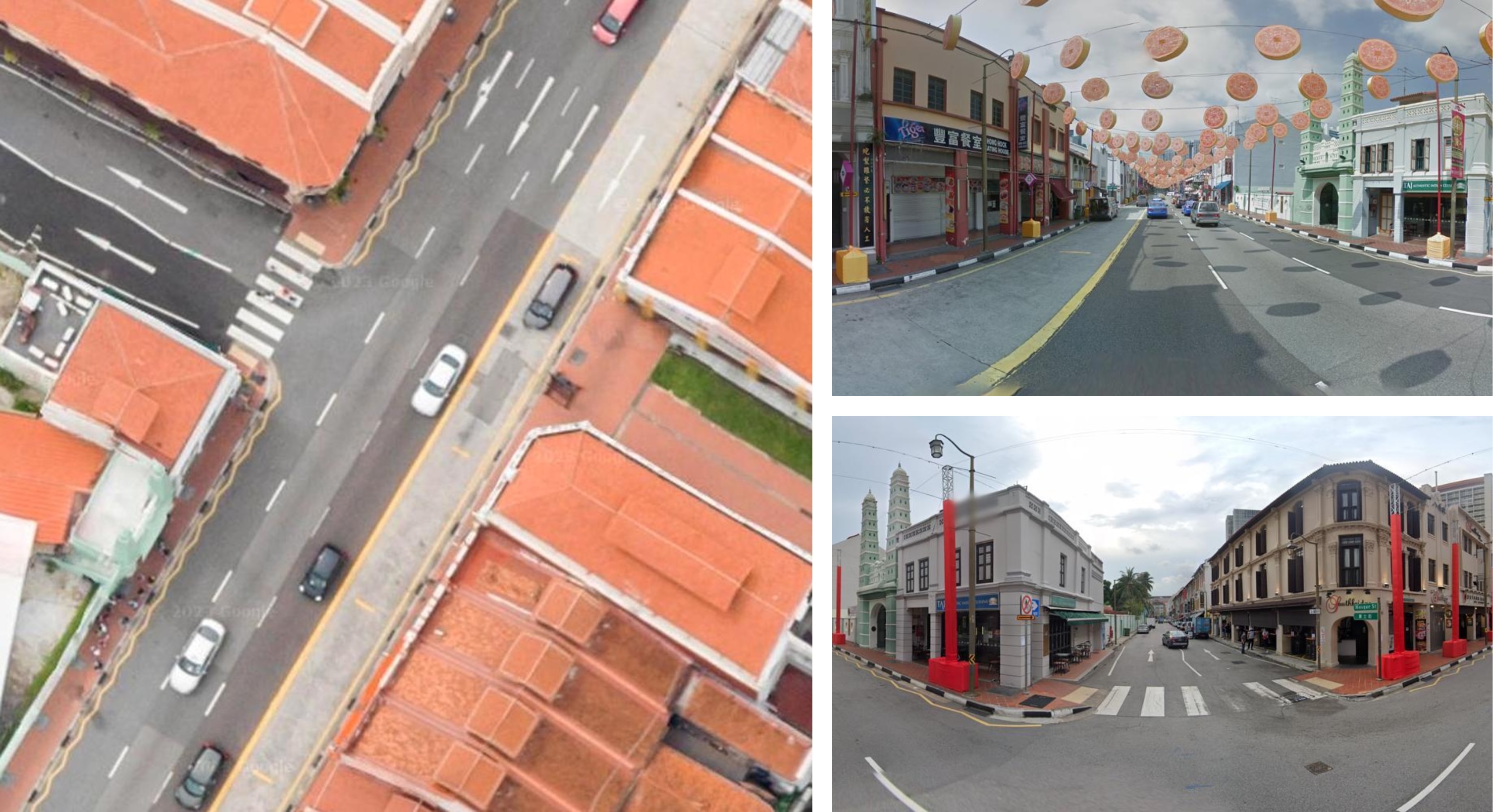}
    \caption{SpaGBOL node data: satellite image \& 2/5 corresponding \ac{fov}-cropped streetview images - shown at different yaws.}
    \label{fig:data_example}
    \vspace{-1.5em}
\end{figure}

\subsection{Evaluation}
\begin{figure*}[h!]
    \centering
    \includegraphics[width=\textwidth]{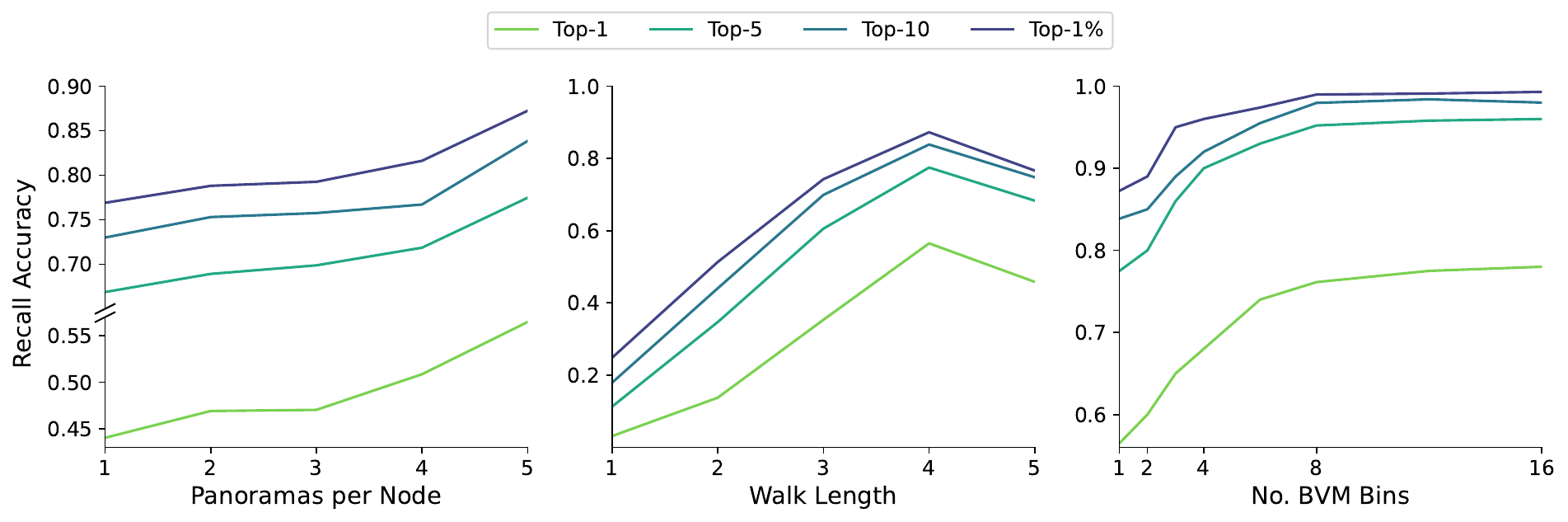}
    \caption{Impact on recall accuracies when \ac{paper_name} characteristics are varied.}
    \label{fig:ablation_graphs}

    \vspace{-1em}
    
\end{figure*}

We evaluate with Top-K recall accuracy, similar to previous works \cite{Zhu2022TransGeoTI}, \cite{shore2023bevcv}, and \cite{sample4geo}, though we enhance performance with retrieval filtering. A query is deemed successful if the correct node is within the Top-K retrievals. Top-K uses the absolute value of K for retrievals whereas Top-K\% uses the K\% length of the database. 
As we are proposing and releasing a novel dataset, we evaluate against previous \ac{cvgl} works whose code is publicly available. We train each approach according to the optimal configurations outlined in their papers/code. As this is the first work to propose graph-based representations and techniques for \ac{cvgl}, we performed a variety of experiments on previous works, aiming to increase fairness. 
One experiment averaged each embedding along sampled walks, another reduced potential reference embeddings at each stage along a walk by performing Top-K retrievals sequentially - aiming to increase retrieval accuracy.
We found empirically that prior works achieve the greatest accuracy when treating each node as an independent retrieval. Thus we train competing techniques in this mode, to provide the strictest baseline possible.
We also evaluate how each technique performs with limited-\ac{fov} images, including those originally designed for panoramic inputs. Table \ref{tab:sota} outlines the performance for each work. \textit{\ac{paper_name}} displays the performance of the network with simple embedding retrieval. \textit{\ac{paper_name}+B} demonstrates how a system can exploit the ability to filter embeddings based on the angles and presence of neighbouring node's edges. For limited-\ac{fov} evaluation, these are only extracted from visible regions of the scene, impacting filtering capabilities. BVM is not utilised where \ac{fov} is below $180\degree$ due to lack of the required visual information. \textit{\ac{paper_name}+YB} improves BVM's potential, displaying the increase in retrieval success when the yaw of the vehicle is also known, i.e. with access to a simple compass. 

\renewcommand{\tabcolsep}{\oldtabcolsep}

\begin{table*}[]
    \centering
    \resizebox{\textwidth}{!}{%
        \begin{tabular}{ccccccccccccc}
            \multicolumn{13}{c}{\textbf{\ac{paper_name}}} \\ \hline
            \multicolumn{1}{r|}{FOV} & \multicolumn{4}{c|}{$360\degree$} & \multicolumn{4}{c|}{$180\degree$} & \multicolumn{4}{c}{$90\degree$} \\ \hline
            \multicolumn{1}{c|}{Model} & Top-1 & Top-5 & Top-10 & \multicolumn{1}{c|}{Top-1\%} & Top-1 & Top-5 & Top-10 & \multicolumn{1}{c|}{Top-1\%} & Top-1 & Top-5 & Top-10 & Top-1\% \\ \hline
            \multicolumn{1}{c|}{CVM \cite{8578856}} & 2.87 & 12.96 & 21.51 & \multicolumn{1}{c|}{28.33} & 2.68 & 9.83 & 15.12 & \multicolumn{1}{c|}{20.23} & 1.02 & 5.87 & 10.15 & 14.81 \\
            \multicolumn{1}{c|}{CVFT \cite{Shi2019OptimalFT}} & 4.02 & 13.02 & 20.29 & \multicolumn{1}{c|}{27.19} & 2.49 & 8.74 & 14.61 & \multicolumn{1}{c|}{19.91} & 1.21 & 5.74 & 10.02 & 13.53 \\
            \multicolumn{1}{c|}{DSM \cite{Shi2020WhereAI}} & 5.82 & 10.21 & 14.13 & \multicolumn{1}{c|}{18.62} & 3.33 & 9.74 & 14.66 & \multicolumn{1}{c|}{21.48} & 1.59 & 5.87 & 10.11 & 16.24 \\
            \multicolumn{1}{c|}{L2LTR \cite{L2LTR}} & 11.23 & 31.27 & 42.50 & \multicolumn{1}{c|}{49.52} & 5.94 & 18.32 & 28.53 & \multicolumn{1}{c|}{35.23} & 6.13 & 18.70 & 27.95 & 34.08 \\
            \multicolumn{1}{c|}{GeoDTR+ \cite{GeoDTR+}} & 17.49 & 40.27 & 52.01 & \multicolumn{1}{c|}{59.41} & 9.06 & 25.46 & 35.67 & \multicolumn{1}{c|}{43.33} & 5.55 & 17.04 & 24.31 & 31.78 \\
            \multicolumn{1}{c|}{SAIG-D \cite{SAIG}} & 25.65 & 51.44 & 62.29 & \multicolumn{1}{c|}{68.22} & 15.12 & 35.55 & 45.63 & \multicolumn{1}{c|}{53.10} & 7.40 & 21.76 & 31.14 & 37.14 \\
            \multicolumn{1}{c|}{Sample4Geo \cite{sample4geo}} & 50.80 & 74.22 & 79.96 & \multicolumn{1}{c|}{82.32} & 37.52 & \textbf{64.52} & 71.92 & \multicolumn{1}{c|}{76.39} & 6.51 & 20.61 & 30.31 & 36.12 \\ \hline
            \multicolumn{1}{c|}{\ac{paper_name}} & \textbf{56.48} & \textbf{77.47} & \textbf{83.85} & \multicolumn{1}{c|}{\textbf{87.24}} & \textbf{40.88} & 63.79 & \textbf{72.88} & \multicolumn{1}{c|}{\textbf{78.28}} & \textbf{18.63} & \textbf{43.20} & \textbf{54.05} & \textbf{61.20} \\
            \multicolumn{1}{c|}{\ac{paper_name}+B} & 64.01 & 86.54 & 92.09 & \multicolumn{1}{c|}{94.64} & 52.01 & 82.20 & 89.47 & \multicolumn{1}{c|}{93.62} & - & - & - & - \\
            \multicolumn{1}{c|}{\ac{paper_name}+YB} & 76.13 & 95.21 & 97.96 & \multicolumn{1}{c|}{98.98} & 66.82 & 92.69 & 96.38 & \multicolumn{1}{c|}{97.30} & - & - & - & - \\ \hline
            \multicolumn{13}{c}{\textbf{VIGOR-Graph}} \\ \hline
            \multicolumn{1}{r|}{FOV} & \multicolumn{4}{c|}{$360\degree$} & \multicolumn{4}{c|}{$180\degree$} & \multicolumn{4}{c}{$90\degree$} \\ \hline
            \multicolumn{1}{c|}{Model} & Top-1 & Top-5 & Top-10 & \multicolumn{1}{c|}{Top-1\%} & Top-1 & Top-5 & Top-10 & \multicolumn{1}{c|}{Top-1\%} & Top-1 & Top-5 & Top-10 & Top-1\% \\ \hline
            \multicolumn{1}{c|}{CVM \cite{8578856}} & 1.83 & 7.80 & 11.90 & \multicolumn{1}{c|}{22.75} & 1.79 & 5.49 & 9.63 & \multicolumn{1}{c|}{16.99} & 1.39 & 4.31 & 8.58 & 15.08 \\
            \multicolumn{1}{c|}{CVFT \cite{Shi2019OptimalFT}} & 5.01 & 12.99 & 18.48 & \multicolumn{1}{c|}{28.93} & 1.96 & 6.28 & 9.89 & \multicolumn{1}{c|}{16.51} & 1.31 & 3.57 & 6.28 & 11.29 \\
            \multicolumn{1}{c|}{DSM \cite{Shi2020WhereAI}} & 6.19 & 16.51 & 22.14 & \multicolumn{1}{c|}{32.64} & 1.05 & 2.31 & 3.70 & \multicolumn{1}{c|}{7.67} & 0.44 & 1.48 & 2.66 & 5.36 \\
            \multicolumn{1}{c|}{L2LTR \cite{L2LTR}} & 6.41 & 17.52 & 26.45 & \multicolumn{1}{c|}{37.91} & 3.09 & 8.37 & 12.20 & \multicolumn{1}{c|}{20.78} & 1.87 & 6.75 & 10.12 & 17.08 \\
            \multicolumn{1}{c|}{GeoDTR+ \cite{GeoDTR+}} & 3.09 & 11.07 & 17.08 & \multicolumn{1}{c|}{28.24} & 2.05 & 6.71 & 11.20 & \multicolumn{1}{c|}{20.22} & 1.48 & 5.19 & 9.37 & 17.43 \\
            \multicolumn{1}{c|}{SAIG-D \cite{SAIG}} & 7.63 & 17.47 & 24.92 & \multicolumn{1}{c|}{36.17} & 5.27 & 14.55 & 21.79 & \multicolumn{1}{c|}{32.81} & 2.88 & 7.97 & 13.16 & 21.00 \\
            \multicolumn{1}{c|}{Sample4Geo \cite{sample4geo}} & \textbf{32.03} & 54.73 & 64.10 & \multicolumn{1}{c|}{75.90} & \textbf{13.92} & 31.07 & 36.17 & \multicolumn{1}{c|}{54.23} & 1.35 & 4.40 & 7.93 & 14.81 \\ \hline
            \multicolumn{1}{c|}{\ac{paper_name}} & 31.88 & \textbf{57.99} & \textbf{67.47} & \multicolumn{1}{c|}{\textbf{77.56}} & 13.36 & \textbf{31.53} & \textbf{41.66} & \multicolumn{1}{c|}{\textbf{54.59}} & \textbf{6.51} & \textbf{18.95} & \textbf{27.07} & \textbf{41.22} \\
            \multicolumn{1}{c|}{\ac{paper_name}+B} & 47.99 & 74.63 & 83.45 & \multicolumn{1}{c|}{91.40} & 19.17 & 42.53 & 52.88 & \multicolumn{1}{c|}{66.25} & - & - & - & - \\
            \multicolumn{1}{c|}{\ac{paper_name}+YB} & 58.21 & 81.49 & 88.69 & \multicolumn{1}{c|}{94.32} & 21.88 & 47.25 & 58.17 & \multicolumn{1}{c|}{69.96} & - & - & - & - \\ \hline
        \end{tabular}
    }
    \caption{Benchmark Dataset Test Recall Accuracies.}
    \label{tab:sota}

    \vspace{-1.25em}

\end{table*}

\begin{table}[t!]
    \centering
    \begin{tabular}{l|cccc}
                                   & \multicolumn{4}{c}{Train}        \\
        \multicolumn{1}{c|}{Model} & Top-1 & Top-5 & Top-10 & Top-1\% \\ \hline
        ConvNeXt-T                 & 52.93 & 70.80 & 88.01  & 92.87   \\
        C+GNN                      & 79.04 & 95.80 & 97.36  & 99.75   \\
        C+G+Bearing                & 84.03 & 97.92 & 99.55  & 99.91   \\
        C+G+B+Yaw                  & 85.89 & 98.82 & 99.29  & 99.97   \\
                                   & \multicolumn{4}{c}{Test}         \\ \hline
        ConvNeXt-T                 & 15.00 & 44.80 & 60.00  & 67.58   \\
        C+GNN                      & 56.48 & 77.47 & 83.85  & 87.24   \\
        C+G+Bearing                & 64.01 & 86.54 & 92.09  & 94.64   \\
        C+G+B+Yaw                  & 76.13 & 95.21 & 97.96  & 98.98  
    \end{tabular}
    \caption{Ablation study demonstrating the performance impact from each component of \ac{paper_name}.}
    \label{tab:abla}
    \vspace{-1.2em}
\end{table}
Results from both datasets show that our proposal achieves significant improvements over previous works, specifically when performing \ac{cvgl} in densely sampled city-scale graphs. 
We demonstrated in Figure \ref{fig:ablation_graphs} that the inclusion of multiple streetview images per node improves generalisation - increasing test performance by approximately 10\% for each metric, when increasing from one streetview image per node to five. 
Also showing that when evaluating with the \ac{paper_name} dataset, the optimal walk length was four, with performance dropping when exceeding this. 
Utilising our GNN-based network achieves performance increases of 11.18\% on Top-1 retrievals on \ac{paper_name}. 
Also showing that the filtered GNN embeddings are more robust to reduced \ac{fov} inputs with our Top-1 relatively decreasing by approximately 12\% compared to previous \ac{sota} performance's reduction of 26\%, when reducing input \ac{fov} to $180\degree$.
Utilising graph characteristics which allow for our bearing filtering proposal, demonstrates that this can achieve relative performance increases beyond our standard retrieval system of $\approx35\%$ when \ac{fov} is $360\degree$, and 67\% when \ac{fov} is $180\degree$.

\subsection{Ablation Study}
To verify components contribute as intended within our proposed system, we display an ablation of constituents in Table \ref{tab:abla}. The base model is only the feature extraction, trained for single-image retrieval as it has no graph walk capability. Adding our GNN greatly improved performance, outputting geo-spatially strong embeddings from the more discriminative network. We then add bearing vector filtering which further boosts performance around 15\% by removing incompatible nodes. Finally, adding the camera yaw to the system optimised performance by filtering with aligned bearing vectors. We determine the optimal walk length of our system with the \ac{paper_name} dataset - varying the walk length of all sampled walks. Visible in Figure \ref{fig:ablation_graphs}, the system's performance dramatically increases when walk lengths are larger than two - with the optimal for this dataset being random walks of length four. To improve generalisation of our network and future works, we include multiple streetview panoramas for each node in the graphs. These images were captured across a period of around a decade - leading to varying content, weather, and lightning.

\section{Conclusion \& Future Work}
\vspace{-0.5em}
In this paper, we successfully progress \ac{cvgl} towards real-world application, demonstrating the benefits of advancing the field from single-image and image-sequence representations towards explicitly structured graphs. We release a comprehensive novel dataset focused on regions most likely to benefit from \ac{cvgl} - dense \ac{gnss}-denied urban regions. We have presented an approach using graph representations and GNNs to significantly aid \ac{cvgl} by exploiting the relationship between image features, their geographic proximity, and geo-spatial structures. Furthermore we have demonstrated how performance may be boosted by implementing \ac{bvm} according to observed road bearings. Evaluating against previous approaches, we increase retrieval performances by more than 11.18\% for Top-1 retrievals - boosting up to 49.86\% when utilising the \ac{bvm} capabilities of graph representation.

\subsection{Future Work}
\vspace{-0.5em}
We have demonstrated the utility of graphs for \ac{cvgl}, effectively verifying various benefits of such approaches. However, there are some limitations that must be addressed in future works.
Although closer to real-world feasibility than prior datasets/techniques, the granularity of our dataset limits precision - only capable of localising to the nearest road junction. 
Within our test set, the median length of edges is 73 metres. 
This could be naively addressed by incorporating additional sensors for localising between nodes, such as using an IMU for measuring between successful retrievals. 
Future works may overcome this obstacle by introducing hierarchical structures such as sub-graph representations for each edge on the corpus graph, allowing for secondary localisation once the nearest node has been determined against the city-scale graph.

\vspace{-0.5em}

\section{Acknowledgements}
\vspace{-0.4em}
This work was partially funded by the EPSRC under grant agreement EP/S035761/1, FlexBot - InnovateUK project 10067785, and the author was financially supported by G-Research.

{\small
\bibliography{packages/egbib}
}

\end{document}